\documentclass{article}

\usepackage[preprint]{neurips_2021}

\usepackage[utf8]{inputenc} % allow utf-8 input
\usepackage[T1]{fontenc}    % use 8-bit T1 fonts
\usepackage{hyperref}       % hyperlinks
\usepackage{url}            % simple URL typesetting
\usepackage{booktabs}       % professional-quality tables
\usepackage{graphicx}
\usepackage{amsfonts}       % blackboard math symbols
\usepackage{nicefrac}       % compact symbols for 1/2, etc.
\usepackage{microtype}      % microtypography
\usepackage{xcolor}         % colors

\title{Automated Detection of Patients in Hospital Video Recordings}

\author{%
  Siddharth Sharma \\
  Stanford University\\
  \texttt{sidshr@stanford.edu} \\
  \And
  Florian Dubost \\
  Stanford Medicine \\
  \texttt{fdubost@stanford.edu} \\
  \AND
  Christopher Lee-Messer \\
  Stanford Neurology \\
  \texttt{cleemess@stanford.edu} \\
  \And
  Daniel Rubin \\
  Stanford Medicine \\
  \texttt{dlrubin@stanford.edu} \\
}

\begin{document}

\maketitle

\begin{abstract}
In a clinical setting, epilepsy patients are monitored via video electroencephalogram (EEG) tests. A video EEG records what the patient experiences on videotape while an EEG device records their brainwaves. Currently, there are no existing automated methods for tracking the patient’s location during a seizure, and video recordings of hospital patients are substantially different from publicly available video benchmark datasets. For example, the camera angle can be unusual, and patients can be partially covered with bedding sheets and electrode sets. Being able to track a patient in real-time with video EEG would be a promising innovation towards improving the quality of healthcare. Specifically, an automated patient detection system could supplement clinical oversight and reduce the resource-intensive efforts of nurses and doctors who need to continuously monitor patients. We evaluate an ImageNet pre-trained Mask R-CNN, a standard deep learning model for object detection, on the task of patient detection using our own curated dataset of 45 videos of hospital patients. The dataset was aggregated and curated for this work. We show that without fine-tuning, ImageNet pre-trained Mask R-CNN models perform poorly on such data. By fine-tuning the models with a subset of our dataset, we observe a substantial improvement in patient detection performance, with a mean average precision of 0.64. We show that the results vary substantially depending on the video clip.
\end{abstract}

\section{Introduction}

Epilepsy is the second most common brain disorder behind migraines. Seizure detection and localization are ongoing fields of study in neurology that aim to create algorithms that can successfully identify epileptic seizures. Seizure patients can be tracked by two mediums: electroencephalographic (EEG) signals and video EEG. An EEG signal is a multi-channel time series indicating the patient’s brain waves while a video EEG is a video monitor of the patient’s physical movement and of its surrounding environment. In the past, advances have been made towards the use of pure EEG signals for automated seizure detection \citep{hussein2018epileptic, saab2020}. However, the automated analysis of video recordings of hospital patients has not been addressed. Analysis of the video could allow detecting events that are not visible on other modalities--such as EEG--and could contain essential information for automated seizure detection. For example, repetitive motions, such as chewing while eating, can trigger an EEG signal that resembles seizure patterns and misleads EEG-based automated detection algorithms.  

In many human detection and tracking tasks, the face of most of the body of the targets is visible \citep{everingham15, shao2018, dollar2011}. Epilepsy patient detection is considerably more difficult because their head is often covered with an EEG headset and their body by blankets or cables.

For this work, we aggregated and curated a dataset of 45 video recordings of hospital patients by leveraging continuous video recordings acquired during clinical routine. We use this dataset to compared multiple standard deep learning detection models for the task of epilepsy patients detection. We fine-tune variants of Mask R-CNN architecture \citep{he2017} and compare to their ImageNet pre-trained non-fined tuned version. We use the mean average precision with a bounding box overlap of at least 50 percent (mAP50) to evaluate the performance of our models in patient detection. We show that fine-tuned models outperform existing state-of-the-art pre-trained person detection models. We believe that our patient detection pipeline has vast applications in the clinic and may serve as a promising step towards highly accurate seizure detection and localization.

\section{Dataset}

\begin{figure}
    \centering
    \includegraphics[width=8cm]{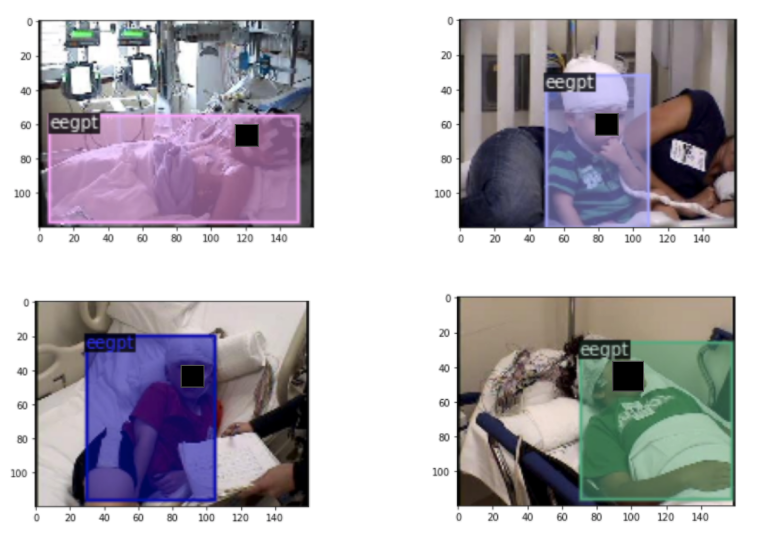}
    \caption{Sample annotated EEG patient video frames. Patient faces are redacted for privacy.}
    \label{fig:examples_annotations}
\end{figure}

We created a dataset for hospital patient video detection using continuous video recordings acquired during the clinical routine. We selected 50 videos clips from different neonates, children, teenagers, and adults patients with a heavier concentration of neonates and children. The videos were acquired under different camera angles, often with partial occlusion or the patient with blankets or machines, sometimes with the patient lying in bed, sometimes with relatives or nurses in the footage. We attempted to have as much variation as possible (Figure \ref{fig:examples_annotations}). 

Each annotated video clip consists of 600 frames acquired at 15 frames per second (40 seconds per clip) with a frame resolution of 240x320. This accounts for 30,000 frames of patient activity for the full dataset. Five videos were dropped during processing due to formatting errors.

The Computer Vision Annotation Tool (CVAT) \citep{cvat} platform was used to create bounding boxes around the patients. We included the body parts occluded by the blankets in our bounding box annotations. For reference, we present four unique, annotated images of epilepsy patients in Figure \ref{fig:examples_annotations}.  

This annotated video dataset was randomly split into 31 videos for training, 11 for validation, and 5 for testing. The training set is used to update the network's parameters, the validation set to evaluate the generalization performance during training, and the test set is left out to perform the final independent evaluation once the model is trained.

\section{Methods}
\subsection{Object Detection}

Convolutional Neural Networks have become the standard for image recognition and computer vision tasks. Tasks like object detection and semantic segmentation have also become crucial due to applications in sectors like autonomous driving and robotics. For the task of seizure patient detection and tracking, we process every video frame individually using a 2D Mask R-CNN \citep{he2017}. Mask R-CNN is a conceptually simple and flexible framework for object instance detection and segmentation. It builds on the work of Faster R-CNN \citep{ren2015}, another detection model, by adding a branch for predicting object masks in parallel to the branch for bounding box recognition. It has delivered top performance on the COCO 2016 dataset, a benchmark for object detection tasks \citep{lin2014}. We utilize both the 50 and 101 layer variants of Mask R-CNN.

\subsection{Mask R-CNN}

 Mask R-CNN is an overall two-stage procedure that uses parallelism in regard to the class and bounding box. Every candidate object has two outputs: a class label and a bounding box offset.  
 
 Building from Fast R-CNN, the first stage of a Regional Proposal Network is adopted. Pixel-to-pixel alignment is added and features are extracted using region of interest pooling (RoIPool) from each candidate box. RoIPool is a technique that performs quantization for the pixels via pooling \citep{he2017}. This is combined with a RoIAlign layer \citep{he2017} whose task is to properly realign the extracted features with the input following the quantization performed by RoIPool.

Mask R-CNN is unique in that it outputs a binary mask for each region of interest in contrast to most object segmentation systems where classification depends on mask predictions. For each region of interest, a multi-task loss is defined as:

\begin{equation}
    L = L_c + L_b + L_m,
\end{equation}

where $L_c$ and $L_b$ are the classification and bounding box losses defined in Fast R-CNN \cite{girshick2015}, and $L_m$ is the loss that generates class-wise masks to prevent competition between classes.

The mask representation is also valuable to the speed performance of Mask R-CNN because it encodes an input object's spatial layout and produces faster inference. 

For its backbone, Mask R-CNN features a convolutional architecture for feature extraction. This architecture is that of a vanilla 2D ResNet \citep{he2016} of either 50 or 101 layers. A Feature Pyramid Network \cite{lin2017} is subsequently added to create levels of prioritization for features. Feature Pyramid Network uses a top-down architecture with lateral connections.

\subsection{Model Fine-tuning}

We fine-tune the 50 and 101 layers ResNet variants of the Mask R-CNN architecture on a training dataset of 30 videos. These networks were pre-trained on ImageNet \citep{deng2009}. To fine-tune this model for the patient detection task, we perform hyperparameter optimization and data augmentation as part of a custom training loop. 

For training, we use stochastic gradient descent. We use the learning rate (0.00001) and batch size (512) that achieve the best performance on the validation set. To improve generalization performance, we also added on the fly data augmentation during training, namely random brightness changes (90 to 110 percent), random flipping in y with a 0.5 probability, resizing frames to 360x480 and applying random crop of 200x300, random rotation from 0 to 15 degrees, and random contrast from 90 to 110 percent.

We train the model for 50,000 iterations of the training set and evaluate the loss function on the validation set every 1,000 iterations. For the final evaluation on the test set, we select the model that minimizes the loss function on the validation set.

\begin{figure*}[t]
  \centering
  \includegraphics[width=\textwidth,height=7cm]{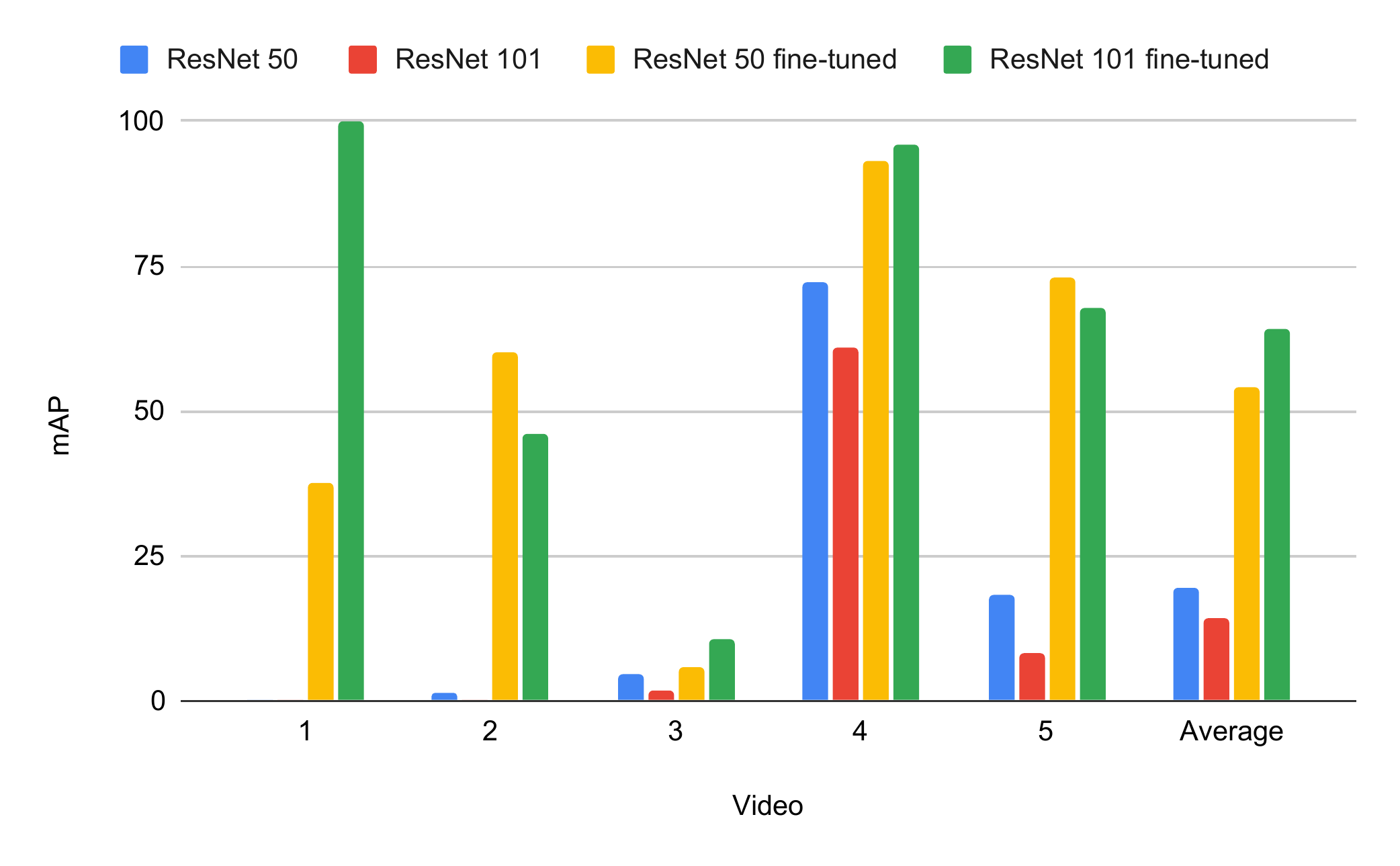}
  \caption{Comparison of fine-tuned models to pre-trained only models with mAP at IoU of 0.5. Results are reported on the test set.}
  \label{fig:results}
\end{figure*}

\subsection{Technical Specfications}
We use Detectron2, a wrapper for object detection with neural networks created by Facebook AI Research \citep{wu2019}. We use Pytorch 1.9.0 and train our models with two NVIDIA TITAN RTX GPUs.

\subsection{Evaluation}

To evaluate the performance of our model, we used the mean average precision (mAP) with a bounding box overlap of at least 50 percent intersection over union (IoU). IoU is measured by calculating the Area of Overlap and dividing it by the Area of Union. 

mAP50 is a common object detection metric and is used for example in the Pascal Visual Object Classes (VOC) Challenge \citep{everingham15}. This metric is best for the task of patient detection since it is more difficult to define the location of patient bounding boxes compared to classical human detection tasks. The mAP50 is computed video-wise and averaged over videos.

\section{Results and Discussion}

The main result we identify in this work is the performance of our custom model (an ImageNet pre-trained Mask R-CNN that was fine-tuned on a training set of 40 patient videos) versus existing state-of-the-art person detection models of the same architecture: ImageNet pre-trained Mask R-CNNs. Both Resnet 101 and 50 variants of Mask R-CNN achieve substantially higher patient detection performance after fine-tuning (Figure \ref{fig:results}). The fine-tuned ResNet 101 variant achieves the best result with a mAP50 of 64.0. The fine-tuned ResNet 50 is second best with a mAP50 of 53.9. Without fine-tuning, and despite ImageNet pretraining, these models reached a much lower patient detection performance: 19.3 of mAP50 for ResNet 50 and 14.1 for ResNet 101, failing to detect the patient in most cases.

As discussed earlier, we assume that detecting epilepsy patients is more difficult due to the presence of obstacles like blankets, nurses, wires, etc. Moreover, while most person detection tasks show the arms and legs of the targets being visible, it is rare to see the full body of a seizure patient. Figure \ref{fig:results} shows that ImageNet pretraining without fine tuning detects patients reasonably well in the absence of bed covers but does not generalizes well to covered patients.

One of the limitations of our work is that we do leverage the temporal relation of frames and process them independently. Temporal information across frames can be crucial to understanding the behaviors of different objects in a scene \citep{carreira2017} such as epilepsy patient tracking. Temporal information could be leveraged using a 3D ResNet as the backbone of the Mask R-CNN. However, depending on the size of the video clips, we be limited by the GPU memory. We could also post-process the predicted bounding boxes coordinates as a second optimization step.

\begin{figure} [t]
    \centering
    \includegraphics[width=8cm]{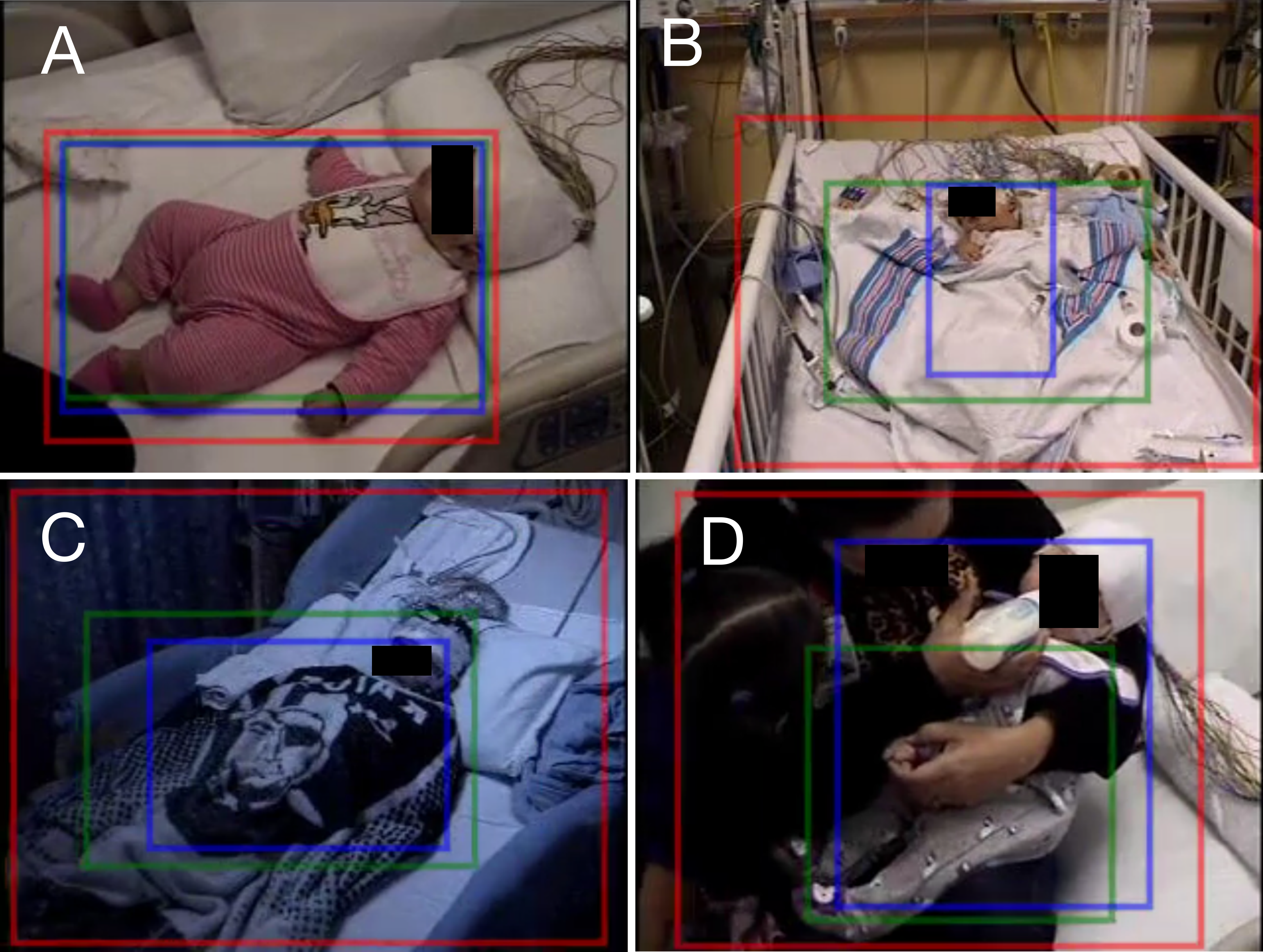}
    \caption{Examples of patient detections. The ground truth bounding box is in blue, the detections of the fine-tuned ResNet 101 Mask R-CNN in green, and those of not ImageNet pre-trained ResNet 50 Mask R-CNN in red. Patient faces are redacted for privacy. Frame A has been extracted from video 4 (in Figure \ref{fig:results}), where all models achieve high mAP50; Frame B from video 3, where all models achieve poor performance; frame C from video 1 where the fine-tuned ResNet 101 Mask R-CNN achieves perfect mAP50, while both baseline ImageNet pre-trained models get a mAP50 of 0; and frame D, from video 5, where we can observe that the fine-tuned ResNet 101 Mask R-CNN may be biased to bed cover or adult hands detection.}
\label{fig:visu}
\end{figure}

\section{Conclusion}

We presented patient detection results using our own dataset of hospital patient video aggregated and curated for this article. We estimated the need of fine-tuning state-of-the-art ImageNet pretrained models to perform correct patient detection. Fine-tuned neural networks could achieve a hospital patient detection performance of 64.0 mean average precision in continuous video recordings. Without fine-tuning, these models reached at best 19.3 of mean average precision and failed to detect the patient in most cases. This underlines the importance of fine-tuning object detection models in hospital settings. Given these promising results pave the way to building an ecosystem for finetuning object detection models in hospital video recordings as a whole. 

\bibliographystyle{plainnat}
\bibliography{jmlr-sample}

\begin{thebibliography}{15}
\providecommand{\natexlab}[1]{#1}
\providecommand{\url}[1]{\texttt{#1}}
\expandafter\ifx\csname urlstyle\endcsname\relax
  \providecommand{\doi}[1]{doi: #1}\else
  \providecommand{\doi}{doi: \begingroup \urlstyle{rm}\Url}\fi

\bibitem[Carreira and Zisserman(2017)]{carreira2017}
Joao Carreira and Andrew Zisserman.
\newblock Quo vadis, action recognition? a new model and the kinetics dataset.
\newblock In \emph{proceedings of the IEEE Conference on Computer Vision and
  Pattern Recognition}, pages 6299--6308, 2017.

\bibitem[Deng et~al.(2009)Deng, Dong, Socher, Li, Li, and Fei-Fei]{deng2009}
Jia Deng, Wei Dong, Richard Socher, Li-Jia Li, Kai Li, and Li~Fei-Fei.
\newblock Imagenet: A large-scale hierarchical image database.
\newblock In \emph{2009 IEEE conference on computer vision and pattern
  recognition}, pages 248--255. Ieee, 2009.

\bibitem[Dollar et~al.(2011)Dollar, Wojek, Schiele, and Perona]{dollar2011}
Piotr Dollar, Christian Wojek, Bernt Schiele, and Pietro Perona.
\newblock Pedestrian detection: An evaluation of the state of the art.
\newblock \emph{IEEE transactions on pattern analysis and machine
  intelligence}, 34\penalty0 (4):\penalty0 743--761, 2011.

\bibitem[Everingham et~al.(2015)Everingham, Eslami, Van~Gool, Williams, Winn,
  and Zisserman]{everingham15}
M.~Everingham, S.~M.~A. Eslami, L.~Van~Gool, C.~K.~I. Williams, J.~Winn, and
  A.~Zisserman.
\newblock The pascal visual object classes challenge: A retrospective.
\newblock \emph{International Journal of Computer Vision}, 111\penalty0
  (1):\penalty0 98--136, January 2015.

\bibitem[Girshick(2015)]{girshick2015}
Ross Girshick.
\newblock Fast r-cnn.
\newblock In \emph{Proceedings of the IEEE international conference on computer
  vision}, pages 1440--1448, 2015.

\bibitem[He et~al.(2016)He, Zhang, Ren, and Sun]{he2016}
Kaiming He, Xiangyu Zhang, Shaoqing Ren, and Jian Sun.
\newblock Deep residual learning for image recognition.
\newblock In \emph{Proceedings of the IEEE conference on computer vision and
  pattern recognition}, pages 770--778, 2016.

\bibitem[He et~al.(2017)He, Gkioxari, Doll{\'a}r, and Girshick]{he2017}
Kaiming He, Georgia Gkioxari, Piotr Doll{\'a}r, and Ross Girshick.
\newblock Mask r-cnn.
\newblock In \emph{Proceedings of the IEEE international conference on computer
  vision}, pages 2961--2969, 2017.

\bibitem[Hussein et~al.(2018)Hussein, Palangi, Ward, and
  Wang]{hussein2018epileptic}
Ramy Hussein, Hamid Palangi, Rabab Ward, and Z.~Jane Wang.
\newblock Epileptic seizure detection: A deep learning approach, 2018.

\bibitem[Lin et~al.(2014)Lin, Maire, Belongie, Hays, Perona, Ramanan,
  Doll{\'a}r, and Zitnick]{lin2014}
Tsung-Yi Lin, Michael Maire, Serge Belongie, James Hays, Pietro Perona, Deva
  Ramanan, Piotr Doll{\'a}r, and C~Lawrence Zitnick.
\newblock Microsoft coco: Common objects in context.
\newblock In \emph{European conference on computer vision}, pages 740--755.
  Springer, 2014.

\bibitem[Lin et~al.(2017)Lin, Doll{\'a}r, Girshick, He, Hariharan, and
  Belongie]{lin2017}
Tsung-Yi Lin, Piotr Doll{\'a}r, Ross Girshick, Kaiming He, Bharath Hariharan,
  and Serge Belongie.
\newblock Feature pyramid networks for object detection.
\newblock In \emph{Proceedings of the IEEE conference on computer vision and
  pattern recognition}, pages 2117--2125, 2017.

\bibitem[Ren et~al.(2015)Ren, He, Girshick, and Sun]{ren2015}
Shaoqing Ren, Kaiming He, Ross Girshick, and Jian Sun.
\newblock Faster r-cnn: Towards real-time object detection with region proposal
  networks.
\newblock \emph{Advances in neural information processing systems},
  28:\penalty0 91--99, 2015.

\bibitem[Saab et~al.(2020)Saab, Dunnmon, R{\'e}, Rubin, and
  Lee-Messer]{saab2020}
Khaled Saab, Jared Dunnmon, Christopher R{\'e}, Daniel Rubin, and Christopher
  Lee-Messer.
\newblock Weak supervision as an efficient approach for automated seizure
  detection in electroencephalography.
\newblock \emph{NPJ digital medicine}, 3\penalty0 (1):\penalty0 1--12, 2020.

\bibitem[Sekachev et~al.(2020)Sekachev, Manovich, Zhiltsov, Zhavoronkov,
  Kalinin, Hoff, TOsmanov, Kruchinin, Zankevich, DmitriySidnev, Markelov,
  Johannes222, Chenuet, a~andre, telenachos, Melnikov, Kim, Ilouz, Glazov,
  Priya4607, Tehrani, Jeong, Skubriev, Yonekura, vugia truong, zliang7,
  lizhming, and Truong]{cvat}
Boris Sekachev, Nikita Manovich, Maxim Zhiltsov, Andrey Zhavoronkov, Dmitry
  Kalinin, Ben Hoff, TOsmanov, Dmitry Kruchinin, Artyom Zankevich,
  DmitriySidnev, Maksim Markelov, Johannes222, Mathis Chenuet, a~andre,
  telenachos, Aleksandr Melnikov, Jijoong Kim, Liron Ilouz, Nikita Glazov,
  Priya4607, Rush Tehrani, Seungwon Jeong, Vladimir Skubriev, Sebastian
  Yonekura, vugia truong, zliang7, lizhming, and Tritin Truong.
\newblock opencv/cvat: v1.1.0, August 2020.
\newblock URL \url{https://doi.org/10.5281/zenodo.4009388}.

\bibitem[Shao et~al.(2018)Shao, Zhao, Li, Xiao, Yu, Zhang, and Sun]{shao2018}
Shuai Shao, Zijian Zhao, Boxun Li, Tete Xiao, Gang Yu, Xiangyu Zhang, and Jian
  Sun.
\newblock Crowdhuman: A benchmark for detecting human in a crowd.
\newblock \emph{arXiv preprint arXiv:1805.00123}, 2018.

\bibitem[Wu et~al.(2019)Wu, Kirillov, Massa, Lo, and Girshick]{wu2019}
Yuxin Wu, Alexander Kirillov, Francisco Massa, Wan-Yen Lo, and Ross Girshick.
\newblock Detectron2.
\newblock \url{https://github.com/facebookresearch/detectron2}, 2019.

\end{thebibliography}

\end{document}